\documentclass[11pt,a4paper]{article}
\usepackage[hyperref]{eacl2021}
\usepackage{times}
\usepackage{latexsym}

\usepackage{graphicx}
\usepackage{amsmath,amssymb}
\usepackage{subcaption}
\usepackage{booktabs}

\usepackage{microtype}

\aclfinalcopy % Uncomment this line for the final submission
\title{Unsupervised Data Augmentation with 
Naive Augmentation and without Unlabeled Data }

\author{David Lowell \\
Northeastern University \\
  {\tt lowell.d@husky.neu.edu} \\\And
  Brian E Howard \\
  Sciome LLC \\
  {\tt brian.howard@sciome.com} \\\AND
    Zachary C. Lipton \\
 Carnegie Mellon University \\
  {\tt zlipton@cmu.edu} \\\And
  Byron C. Wallace \\
Northeastern University \\
  {\tt b.wallace@northeastern.edu} \\}

\date{}

\begin{document}
\maketitle
\begin{abstract}
Unsupervised Data Augmentation (UDA) is a semi-supervised technique 
that applies a \emph{consistency loss} to penalize
differences between a model's predictions on 
(a) observed (unlabeled) examples; 
and (b) corresponding `noised' examples produced via data augmentation. 
While UDA has gained popularity for text classification,
open questions linger over which design decisions are necessary
and over how to extend the method to sequence labeling tasks.
% 
% This method has recently gained traction for text classification.
% 
In this paper, we re-examine UDA 
and demonstrate its efficacy 
on several sequential tasks.
Our main contribution is an empirical study of UDA
to establish which components of the algorithm
confer benefits in NLP. 
Notably, although prior work has emphasized the use 
of clever augmentation techniques including back-translation,
we find that enforcing consistency between predictions
assigned to observed and \emph{randomly} substituted words
often yields comparable (or greater) benefits 
compared to these complex perturbation models.
Furthermore, we find that applying its consistency loss 
affords meaningful gains \emph{without any unlabeled data at all}, 
i.e., in a standard supervised setting.
In short: UDA need not be unsupervised, 
and does not require complex data augmentation to be effective.
\end{abstract}

\section{Introduction}
%Semi-supervised learning is a core component of modern NLP models: Large transformers pre-trained on general web corpora have afforded considerable progress across a range of tasks \citep{devlin2018bert, lewis2019bart, brown2020language}.
%Pretraining exploits data that is `out of domain' 

%Typically, these transformers are trained as language models on large bodies of out domain text.
%Given the success of this approach, it is reasonable to investigate alternative methods of exploiting the large body of readily available unlabeled language data.

While the advent of large neural models has led to rapid progress 
on a wide spectrum of prediction benchmarks in NLP,
these methods tend to require large amounts of training data. 
This limitation is particularly acute in domains 
such as information extraction from scientific documents, 
where unlabeled in-domain data is plentiful but labeled data is rare 
and requires significant annotator experience to produce.
The cost of acquiring data in such domains has spurred significant interest 
in developing models that can achieve greater extraction accuracy, 
even when the available labelled corpora are small 
\citep{DBLP:journals/corr/abs-1806-04185, maharanapragmatic}. 

In this paper, we investigate unsupervised data augmentation (UDA) \citep{xie2019unsupervised}, 
a recently proposed semi-supervised learning method 
in which models are trained on both labeled and unlabeled in-domain data.
The learning objective for the unlabeled component is to minimize 
the divergence between the model's outputs on a given example 
and its outputs on a perturbed version of the same example.
While this combination of data augmentation with a consistency loss
was previously proposed as \emph{Invariant Representation Learning}
and demonstrated beneficial in speech recognition by \citep{liang2018learning},
UDA applies the method in a semisupervised setting,
in a manner similar to virtual adversarial training \citep{miyato2018virtual}
and the authors demonstrated benefits on both 
computer vision and Natural Language Processing (NLP) tasks.

%created using data augmentation techniques.

%To produce novel examples that are similar to observed examples, one uses data augmentation techniques.
Producing such perturbed examples requires 
% defining
% data augmentation techniques.
specifying
a data augmentation pipeline.
Typically, these apply one or more transformations 
that (hopefully) 
% do not
tend not to
% alter the original examples true label 
alter the applicable label
\citep{Goodfellow:2016:DL:3086952}.
In computer vision, a number of straightforward
and demonstrably effective data augmentation techniques, 
such as image flipping, cropping, rotating,
and various perturbations to the color spectrum
have gained widespread adoption
\citep{huang2016densely, DBLP:journals/corr/ZagoruykoK16}.
More recently, these methods, among others, 
have been successfully applied in concert with UDA
to improve performance on image classification tasks \citep{xie2019unsupervised}.

By contrast, in natural language processing,
there is less consensus about which perturbation models
% data there do not exist simple mathematical augmentations 
can be applied with confidence that they will
not change the applicability of the original label. 
To apply UDA in NLP, researchers have primarily 
focused on \emph{back-translation} 
\citep{sennrich-etal-2016-improving, edunov-etal-2018-understanding},
generating paraphrases by applying a machine translation model
to map a document into a \emph{pivot} language 
and then back into the original language.
% as a means of perturbation.
% This %process is slow and  % bcw: we repeat directly below
In practice, this process produces augmentations of varying quality.
% depending upon quality of the translation models used.
% 
Another problem is that back-translation is slow,
and performance may depend on the choice of translation model.
Incorporating large quantities of unlabeled data 
in the training process is also computationally expensive. % bcw: possibly cite GREEN AI stuff / elaborate here
Given these limitations---and to better characterize why and when UDA helps---we 
investigate whether the benefits of UDA on NLP tasks 
can be achieved using less unlabeled data and/or simpler input perturbations.
To this end, we investigate uniform random word replacement as an augmentation method.
%This method is considered in passing by \citet{xie2019unsupervised}, but is not fully explored.
Random substitution for augmentation has been considered previously 
in the context of translation \cite{wang-etal-2018-switchout}, 
and results using random replacement have recently been investigated 
in the UDA paper \cite{xie2019unsupervised} 
for a single dataset on which it slightly underperforms back-translation. 
Here we deepen this analysis, showing that, surprisingly, random replacement 
is generally competitive with back-translation.
%, and extend it to sequence tagging.
%Surprisingly, we find that random word replacement is as effective as back-translation for the purposes of consistency training. 
Further, we find that significant increases in performance 
can be achieved by applying consistency loss on only small labeled data sets
(although large volumes of in-domain unlabeled data provides further gains).

As an additional contribution, we adapt UDA 
to sequence tagging tasks, which are common in NLP. 
Back-translation is ill-suited to such tasks, 
because we lack alignment between spans of interest 
in the original text and the back-translated paraphrase.
For these problems, we propose and evaluate 
word replacement augmentation strategies for sequence tagging. 
Interestingly, we observe that augmentation via uniform random word replacement yields improvements, 
but that it is more effective to employ a masked language model 
to predict `reasonable' replacements for each word to be replaced.

%Further, back-translation is ill-suited for sequence tagging tasks, as there is no guarantee of alignment between spans of interest in the original document and the back-translated paraphrase.
%We adapt UDA for use in sequence tagging tasks, finding that word replacement is an effective augmentation method in this context as well.
%While we observe that uniform random word replacement is effective, we find that it is more effective to rely on a masked language model to predict `reasonable' replacements in a given masked sequence.
%a sequence labeling context.

\section{Unsupervised Data Augmentation}
\label{section:UDA} 

UDA is a semi-supervised method in which a model is trained---in 
addition to the standard objectives on the labeled data---to 
make similar predictions for an observed example and 
a corresponding perturbed instance, produced via some data augmentation technique.
% The two components of UDA are
Applying UDA requires specifying 
both (i) a consistency loss to be applied on a (original, augmented) example pair;
and (ii) a data augmentation technique to produce the perturbed examples in the first place.

\subsection{Consistency Loss}

As originally proposed by \citet{xie2019unsupervised},
UDA's loss function is a sum over a supervised component and an unsupervised component.
Assuming cross-entropy loss for the former, the loss function is: 
%With standard cross-entropy loss as the supervised component, the training objective is to minimize

% bcw: you had expectations here before, but i think that conflicts with having (x,y) \in L on the bottom (implying ERM) -- the expectation is over a distribution, but this is a specific set that we are minimizing w/r/t
\begin{equation}
    %\mathop{\mathbb{E}}_{x,y \in L}[-\log(\hat{p}(y|x))]+
    %\lambda\mathop{\mathbb{E}}_{x \in \mathcal{U}}[
    %\mathcal{L}(x)
    %]
    \mathop{\sum}_{x,y \in L}[-\log(\hat{p}(y|x))]+
    \lambda\mathop{\sum}_{x \in \mathcal{U}}[
    \mathcal{L}(x)]
\end{equation}

\noindent where $L$ is the set of labeled data,
$\mathcal{U}$ is the set of unlabeled data, 
and $\lambda$ weights the relative contribution 
of the unlabeled term to the total loss. % is a weighting factor. 
%In the context of classification tasks, 
The consistency loss $\mathcal{L}$ is defined 
as the KL-Divergence between model predictions 
for the original and augmented examples.

\begin{equation}
\mathcal{L}(x) = \mathcal{D}_{\text{KL}}\{\hat{p}(y|x)||\hat{p}(y|q(x))\}
\end{equation}

\noindent where $q$ is a data perturbation operation
and $\hat{p}(y|x)$ is the probability distribution 
% over labels induced by the model for $x$.
over labels output by the model given input $x$.

% This definition of $\mathcal{L}$ is not suitable
For sequence tagging tasks
where examples correspond to multiple labels, %a single example is associated with multiple labels.%many constituent labels.
%We adapt UDA to sequence labeling by defining 
we define the consistency loss 
% for sequence labeling 
as the average KL-Divergence 
between per-word model predictions for the original and augmented examples. 
Specifically, we replace the consistency loss above with
% 
% bcw: two things -- (1) n was not defined before, but I infer this is the length of a given sequence and have clarified; (2) if so, we had an off by one error since we started at 0 -- have moved to 1-indexing. but that aside do we really always incur loss for *all* tokens in a given sequence? 
% 
\begin{equation}
    \mathcal{L}(x) = \frac{\sum_{j=1}^{n}\mathcal{D}_{\text{KL}}(\hat{p}(y_j|x)||\hat{p}(y_j|q(x)))}{n}.
    \label{eq:cs-eq}
\end{equation}
% 
% bcw -- before we noted that we denoted word $j$ of $x$ by $x_j$ but i dont think we actually use this anywhere? since y_j depends on all x (for all j), at least as defined above (and i assume this is true)
Here, $n$ denotes sequence length and $\hat{p}(y_j|x)$
are the predicted probabilities assigned by our model 
to labels corresponding to word $j$ of sentence $x$.% ($x_j$). 

\subsection{Data Augmentation Strategies for Text}

%The above consistency loss definitions 
As defined above, consistency loss (for both classification and sequence tagging) 
requires specifying a data perturbation operation $q$ 
that can be applied to observed instances, 
yielding a new, but similar example. %that can be used to augment our dataset.
%Typically, $q$ has been defined as a data augmentation transformation.  % bcw: I found this confusing; not sure what we have in mind where perturbation != augmentation?
If we assume that $q$ %is designed to 
transforms $x$ such that $q(x)$ and $x$ share the same true label, 
then it is clearly desirable that a model would make similar predictions
for $x$ and $q(x)$ (as encouraged by consistency loss).
%This makes data augmentations an intuitive choice for $q$ in UDA. % bcw: again didn't get this discussion really; maybe you're getting at 'being clever' vs not -- but i think this can be stated more explicitly?

However, there is a trade-off between the diversity of instances produced by $q$ for $x$ 
and the likelihood that these will share the same label as $x$.  % bcw: should probably clarify what we mean by 'validty' here (addendum: rewrote a bit)
For example, consider the strategy of paraphrasing via backtranslation. %paraphrases produced by $q$.
A valid paraphrase is one that, with high probability, 
shares the ground-truth label of the input; 
a diverse paraphrase is one that substantially differs from the original. % bcw: should we really talk about two things being 'diverse'? seems like this reuqires >2
%We note that 
\citet{xie2019unsupervised} observed that diversity is more important 
than validity when applying UDA to text classification.
This suggests that it may be possible to effectively use 
an alternative augmentation strategy 
that prioritizes diversity and simplicity at the expense of validity.

\paragraph{Uniform Random Word Replacement}

We propose a variant of $q$ that performs 
a simple uniform random word replacement operation.
Specifically, we define $q(x)$ such that 
most of the time it copies directly from $x$,
but with some probability $p$ it replaces 
each $x_j$ in $x$ with some other word $x_j^\prime$ 
drawn at random from the vocabulary of words 
that appear in $\mathcal{U} \cup L$. Formally:
\begin{equation}
q(x)_j=
\begin{cases}
x_j^\prime & \text{with probability $p$}\\
x_j  & \text{with probability $1-p$}\\
\end{cases}
\end{equation} 

%Each word $x_j^\prime$ is selected uniformly at random from the vocabulary of words appearing in $\mathcal{U} \cup \mathcal{L}$.

This method is simple (and does not require a learned language model), but naive. 
It produces output which is diverse,
but not necessarily valid or even grammatical.
We compare this technique to two cleverer 
model-based data augmentation techniques: 
One for text classification (proposed in prior work) 
and one suitable for sequence tagging (which we introduce here).

\paragraph{Back-Translation}
We use the back-translation machinery described by \citet{xie2019unsupervised}.
Specifically, this entails use of WMT’14 English-French \citep{bojar-EtAl:2014:W14-33} 
translation models in both directions with random sampling 
with a tunable temperature in place of beam search for generation. % bcw: if we're going to incldue this remark, probably good to clarify a bit? probably most readers will not be immediately familiar with this decoding strategy?
We set our temperature to $0.9$, %one of the options 
one of the recommended settings in prior work \cite{xie2019unsupervised}.

\paragraph{Masked Language Model}
Back-translation is not suitable for use in UDA for sequence tagging tasks, 
as in these labels apply to \emph{tokens} and it is not obvious 
how to align tokens in a given paraphrase with those in the original text.
%In the sequence tagging domain, 
More specifically, as we have defined it for sequence tagging (Equation \ref{eq:cs-eq}), 
consistency loss penalizes dissimilarity between model predictions 
$p(y_j|x)$ and $p(y_j|q(x))$ for all indices $j$.
However, when $q$ is defined as a back-translation process,
there is no expectation that $x$'s and $q(x)$'s ground-truth labeling will be aligned.
They may not even be of the same length.
Therefore, we instead consider word replacement strategies (at each index $j$),
including (i.i.d.) random replacement, and a model-based word replacement strategy 
that attempts to ensure that the ground-truth labels of $x$ and $q(x)$ are aligned.
Both of these involve individual word substitutions, so $x$ and $q(x)$ will have the same length.

%As in the above uniform random replacement method, 
For the model-based replacement strategy, we again define $q$ such that 
it replaces a given word in $x$ with probability $p$ (otherwise copying from $x$).
%we define $q(x)$ such that it most often copies from $x$, but with probability $p$ it replaces each word $x_j$ with some other word $x_j^\prime$ with probability $p$.
However, here we select $x_j^\prime$ using a masked language model.
%Specifically, to select $x_j^\prime$, 
Specifically, we mask $x_j$ and use BERT \citep{devlin2018bert} 
to induce a probability distribution over all possible words (in its vocabularly) that might appear at position $j$. % bcw: one point of confusion here is that BERT does subwords; how do we handle? might clarify in a footnote for astute reader
We then draw $x_j^\prime$ from the ten most probable words (excluding the original word $x_j$)
with probabilities proportional to the likelihood assigned to these words by BERT.
%BERT's (renormalized) predictions as the probability distribution. % bcw: might want to explain why we do this instead of just drawing from the full distro?
We hypothesize that this method will provide a %significantly 
substantially greater expectation of validity than random replacement,
%This hypothesis is predicated 
on the assumption that BERT is sensitive enough to context 
that it is likely to replace words of one category with other words of the same category. 
% and that the true labels of unreplaced words are unlikely to be altered by such changes in their context.

\begin{table*}[htb]
\small
\centering
\setlength{\tabcolsep}{2pt}
\begin{tabular}{lccccccc}
&ORG & & PER & PER & & PER\\
\midrule
&aberdeen & manager &  roy & aitken & said: 
``it's unfortunate for us that & antoine & cannot play...  \\
\midrule
 &rangers && glen & mc & & he\\
Rand &delegates && cancer & peripheral & & 51,000 \\
\end{tabular}
\caption{Example replacements selected for entity tokens in a randomly selected sentence from CoNLL-2003 using different selection methods. In training, all words are equally likely to be selected for replacement. 
We focus on named entities in this example for illustrative purposes.}
\label{tab:example} 
\vspace{-1.5em}
\end{table*}

%Table \ref{tab:example} compares BERT-based data augmentation to random replacement by showing sample replacements for the entities in a randomly selected sentence from CoNLL.
In Table \ref{tab:example} we show examples of random and BERT-based replacement of entities, in randomly selected sentence from CoNLL.
As expected, random selection does not respect grammaticality or entity classes.% of the original. 
BERT performs better, albeit imperfectly. 
For example, ``aitken'' is replaced with the first token of a surname, ``mc'', 
rather than a full surname and ``antoine'' is replaced with ``he''. 
While the latter substitution is grammatical and semantically similar to the original,
``he'' is not considered a named entity in CoNLL. %the context of CoNLL.

\section{Experimental Setup}

We evaluate our proposed training method on four text classification and three sequence tagging datasets.
Of these, the classification datasets include 
three benchmark sentiment sets (IMDB, Yelp, and Amazon),
and one scientific classification task (evidence inference).
The sequence tagging datasets include one standard NER benchmark dataset (CoNLL-2003) 
and two scientific sequence labeling tasks (EBM-NLP and TAC).
The scientific tasks are of particular interest for this work 
because it is expensive to collect annotations in these specialized domains.

\paragraph{IMDB} \citep{maas-EtAl:2011:ACL-HLT2011} is a sentiment classification dataset 
consisting of movie reviews ($25,000$ in the train set, $25,000$ in the test set) drawn from the IMDB website.
Reviews with a score $\leq 4$ out 10 are considered negative, 
those with scores $\geq 7$ out of 10 are considered positive. 
Neutral reviews are not included.

\paragraph{Yelp} \citep{zhang2015characterlevel} is a sentiment classification dataset 
comprising reviews drawn from Yelp ($560,000$ in the train set, $38,000$ in the test set). 
One and two star reviews are considered negative.
Three and four star reviews are considered positive.

\paragraph{Amazon} \citep{zhang2015characterlevel, 10.1145/2507157.2507163} is a sentiment classification dataset 
consisting of Amazon reviews ($3,600,000$ in the train set, $400,000$ in the test set). 
One and two star reviews are considered negative. 
Four and five star reviews are considered positive. 
Three star reviews are not included.

\paragraph{Evidence Inference} We construct a classification dataset derived from the Evidence Inference dataset \citep{lehman2019inferring,deyoung2020evidence}, a biomedical corpus in which the task is to infer the effect of an intervention on an outcome from an article describing a randomized controlled trial.
The classes correspond to the intervention leading to a \emph{significant increase}, \emph{significant decrease}, or \emph{no significant change} in outcome.
In the original task, the model must first extract relevant evidence sentences from the full text article, and then make a prediction based on this.
We evaluate in the `oracle' setting, in which the model must only classify \emph{given} relevant evidence sentences (${\sim}17,000$ train examples, and ${\sim}2,000$ instances in the test set).

\paragraph{CoNLL-2003} \citep{tjongkimsang2003conll} 
is an NER dataset consisting of annotated Reuters news articles (${\sim}14,000$ sentences in the training set, ${\sim}3,000$ in the test set), labeled with entity categories 
\emph{person}, \emph{organization}, \emph{location}, and \emph{miscellaneous}.

\paragraph{TAC} 
\citep{schmittoverview} comprises annotated ``materials and methods'' sections 
from PubMed Central articles (${\sim}5,500$ sentences in the training set, ${\sim}6,500$ in the test set). Labels are available for 24 entity classes, 
of which we consider the two best represented: \emph{end point} and \emph{test article}.

\paragraph{EBMNLP} 
EBMNLP \citep{DBLP:journals/corr/abs-1806-04185} 
is a corpus of annotated abstracts 
drawn from medical articles describing medical randomized controlled trials (${\sim}28,000$ sentences in the training set, ${\sim}2,000$ in the test set). 
Spans are tagged as describing the 
\emph{patient population}, the \emph{intervention} studied, 
and the \emph{outcome} measured in the trial being described.

For each dataset we simulate a pool of unlabeled data by hiding the annotations of the training set.
We then create five distinct sets of labeled data (ten for sequence tagging tasks) by revealing the annotations for a random subset of the pool, 
forming a labeled training set $L$ and an unlabeled training set $\mathcal{U}$.
For classification tasks, we sample ten examples per class to form $L$, while for CoNLL and EBMNLP, we sample two hundred examples.
For the smaller TAC dataset, we use one hundred.

\paragraph{Training details} For each $L$, 
we then train a model both using only standard supervised learning over $L$ 
and with an additional consistency loss, 
using either uniform random word replacement 
or more complex data augmentation technique 
(backtranslation for classification; BERT-based replacement for sequence tagging).
When training with consistency loss, we evaluate variants in which %the consistency loss is 
we apply this to both $L$ and $\mathcal{U}$, and where we apply it only over $L$.
The latter corresponds to a standard supervised setting (with an additional loss term).
%In the latter case, $\mathcal{U}$ is discarded and the model is trained in a strictly supervised setting

BERT's pretraining task already incorporates unsupervised data \citep{devlin2018bert}.
We therefore also repeat the above experiments using finetuned  weights instead of off-the-shelf pretrained weights. 
The finetuned weights are produced by training 
on BERT's masked language model task 
with $L \cup \mathcal{U}$ as the training data. % bcw: Not clear to me here if we are fine-tuning *starting from* BERT check point or from scratch

In our classification experiments, we use a linear model on top of BERT \citep{devlin2018bert} as a classifier.
%In our sequence tagging experiments,
For sequence tagging, we follow the architecture and hyperparameter choices in prior work \citet{beltagy2019scibert}, which added a conditional random field \citep{lafferty2001conditional} on top of BERT representations.
%This modifies BERT's \citep{devlin2018bert} NER architecture with the addition of a conditional random field \citep{lafferty2001conditional}.
We train all models using Adam \citep{DBLP:journals/corr/KingmaB14}
with a learning rate of $2\mathrm{e}$-05 for classification and $1\mathrm{e}$-3 for sequence tagging.

%Exploratory experiments 
In exploratory experiments we observed that model performance is relatively robust to the choice of $\lambda$ (the weight assigned to the consistency loss term) when large quantities of unlabeled data are available.
We therefore set $\lambda$ to $1$ in all of our semi-supervised experiments.
In %the context of our 
out supervised learning only experiments, we found it necessary to compensate for the lack of unlabeled examples and the corresponding change in the relative weightings of standard and consistency losses.
%We find that t
This can be effectively done by repeating labeled examples, imposing only the consistency loss for them, as though they were unlabeled.
We use a ratio of $20$ ``unlabeled" examples to $1$ labeled example in our supervised experiments. 

% bcw: how did we pick $p$? did we explore sensitivity at all? seems kinda high?
In our augmentation procedure, we set $p$ to $30\%$. 
For biomedical tasks (Evidence Inference, EBMNLP, TAC) model weights are initialized using SciBERT, a model pretrained over scientific papers \citep{beltagy2019scibert}.
For all other tasks,  we initialized parameters to the pretrained $\text{BERT}_\text{BASE}$ weights \citep{devlin2018bert}.
$\text{BERT}_\text{BASE}$ weights are used for CoNLL, and SciBERT weights are used for EBMNLP and TAC when using BERT as a masked-language model for data augmentation.

\section{Results}

\subsection{Classification}

\begin{figure*}[htb]
    \captionsetup[subfigure]{justification=centering}
    \centering
    \begin{subfigure}[]{0.24\linewidth}
		\includegraphics[width=\linewidth]{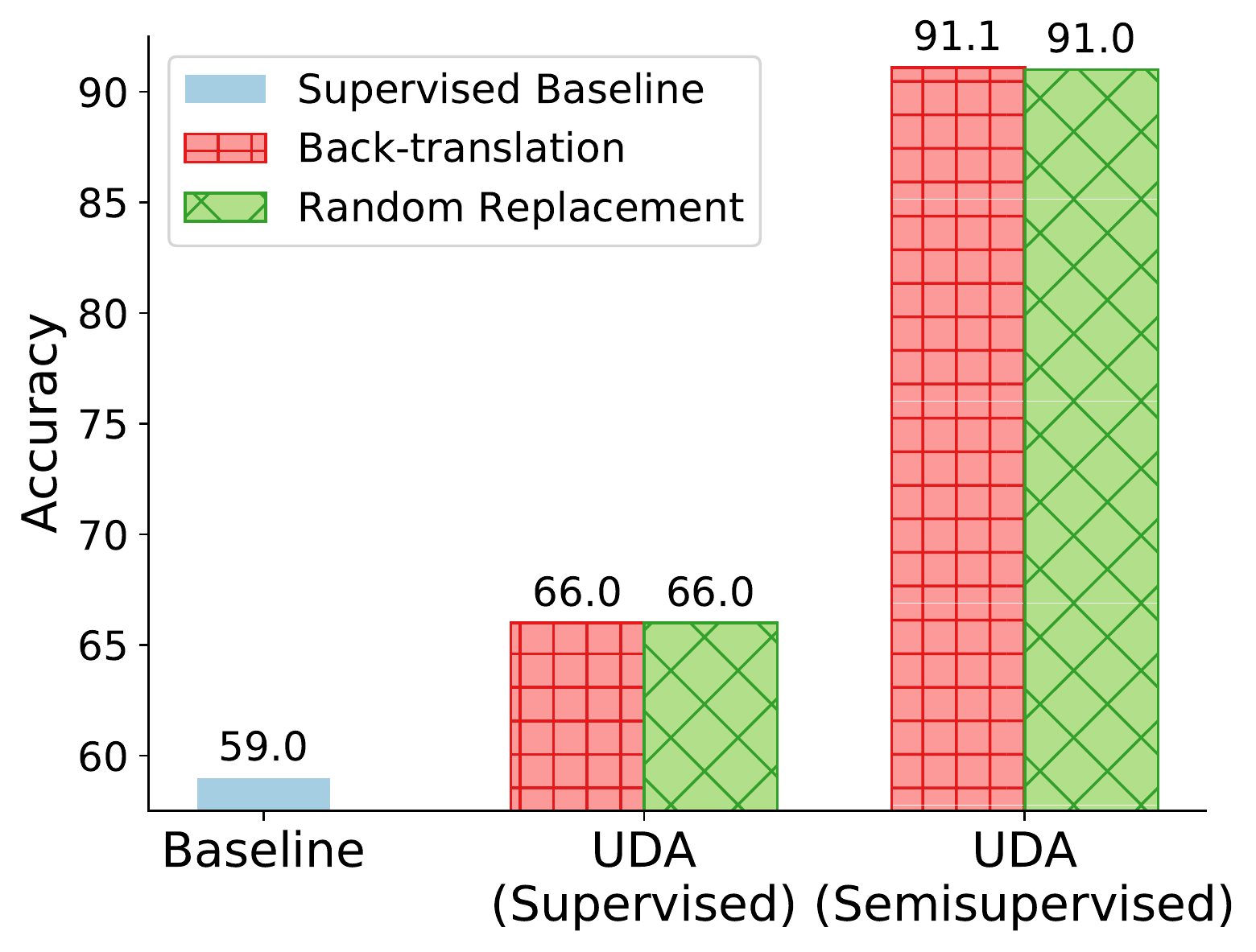}
		\caption{IMDB}
	\end{subfigure}
	\begin{subfigure}[]{0.24\linewidth}
		\includegraphics[width=\linewidth]{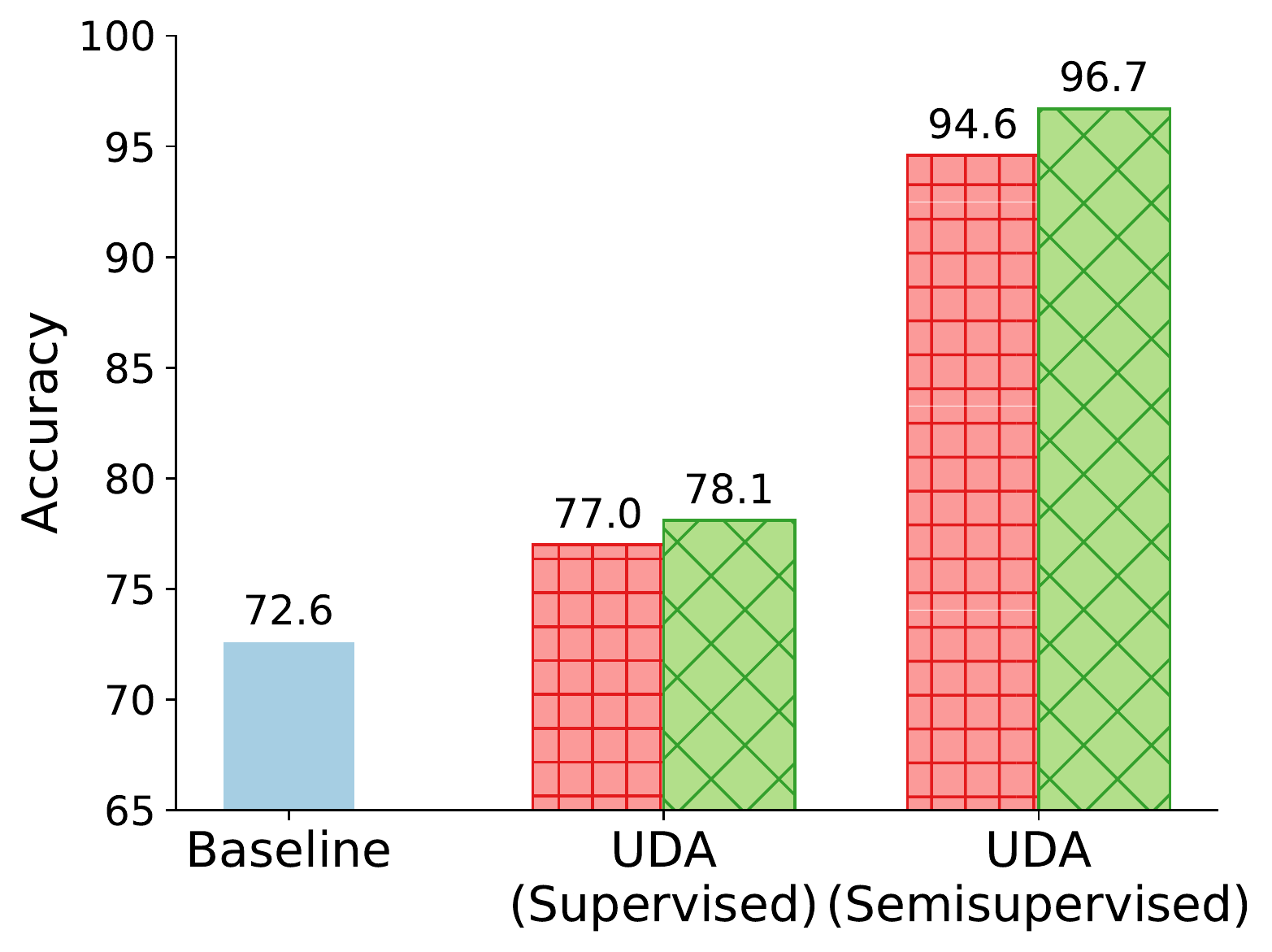}
		\caption{Yelp}
	\end{subfigure}
    \begin{subfigure}[]{0.24\linewidth}
		\includegraphics[width=\linewidth]{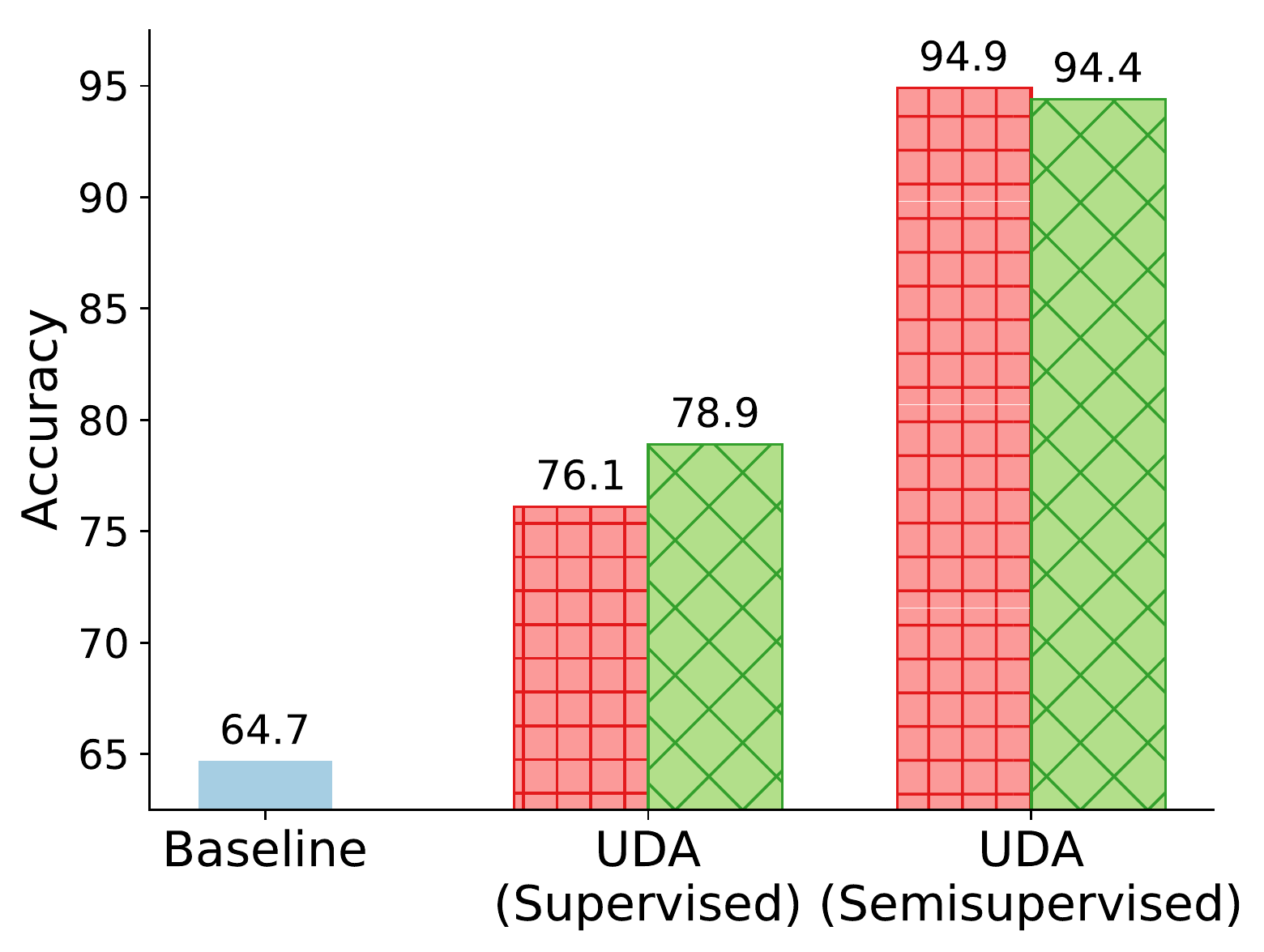}
		\caption{Amazon}
	\end{subfigure}
    \begin{subfigure}[]{0.24\linewidth}
		\includegraphics[width=\linewidth]{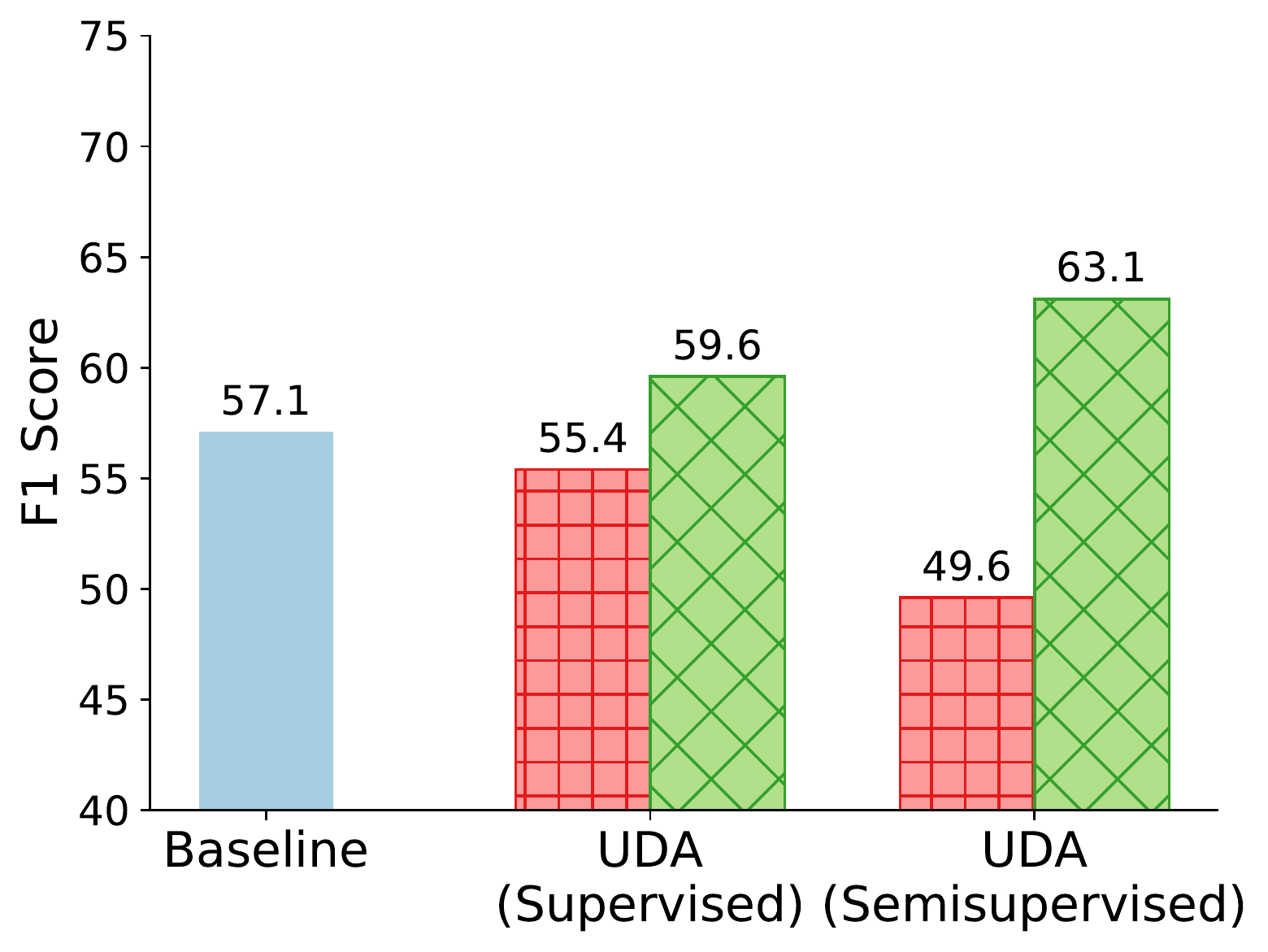}
		\caption{Evidence Inference}
	\end{subfigure}
\caption{Comparison of performance achieved on classification tasks using different variants of UDA. Each bar represents the average performance across five sets of labeled data (labeled data quantity is noted parenthetically). The supervised baseline represents standard ML on the supervised data set only, without any consistency loss. Supervised with consistency loss represents use of consistency loss, but only over the labeled data, with the unlabeled data discarded. Semisupervised with consistency loss represents use of consistency loss over the entire dataset, both labeled and unlabeled.
} % bcw: " Each bar represents the average performance across five sets of labeled data (labeled data quantity is noted parenthetically" ?? is it? we note supervised vs semisupervised parenthetically but not the quantity (right?)
\label{fig:cl_results}
%\vspace{-1.5em}
%\vspace{-.5em}
\end{figure*}

Figure \ref{fig:cl_results} presents the results of our experiments for the classification tasks. 
Both back-translation and random replacement perform well on the IMDB, Yelp, and Amazon datasets. 
Notably, \emph{random replacement consistently achieves results equivalent to or better than those attained with the more computationally complex back-translation method}.

% bcw: it'd be nice to take a look at some of the backtranslated evidence inference examples; i bet the the directionality is broken? 
On the Evidence Inference task, UDA with back-translation under-performs the supervised baseline, with a loss of $7.5$ F1 when the full unlabeled dataset is used.%employed.
This is perhaps unsurprising, given that the models used for back-translation were not trained on scientific text.
%, such as that appearing in the Evidence Inference dataset.
These results suggest that the effectiveness of back-translation is contingent upon the domain similarity of the back-translation model's training data and that of the downstream task.
By contrast, UDA with random replacement produces a modest but meaningful gain over the supervised baseline: $2.5$ F1 with only the labeled data and $6$ F1 with the full unlabeled dataset.

Across all classification experiments---excepting Evidence Inference using back-translation---applying consistency loss to only the labeled data yields improvements over the supervised baseline, albeit less than what is achieved using the full amount of unsupervised data.
Further, these gains are disproportionate to the quantity of data used to attain them. % bcw: I'm not sure I follow what we mean by this?
In the worst case (Yelp with back-translation), using only the supervised data results in only $20\%$ of the potential performance improvement that could be attained using the full set of unlabeled data.
In the best case (Amazon with random replacement), $48\%$ of the potential gain can be achieved without using any unsupervised data at all, despite the fact that the labeled Amazon dataset represents less than $0.001\%$ of the full set.

\begin{figure*}[htb]
    \centering
	\begin{subfigure}[]{0.24\linewidth}
		\includegraphics[width=\linewidth]{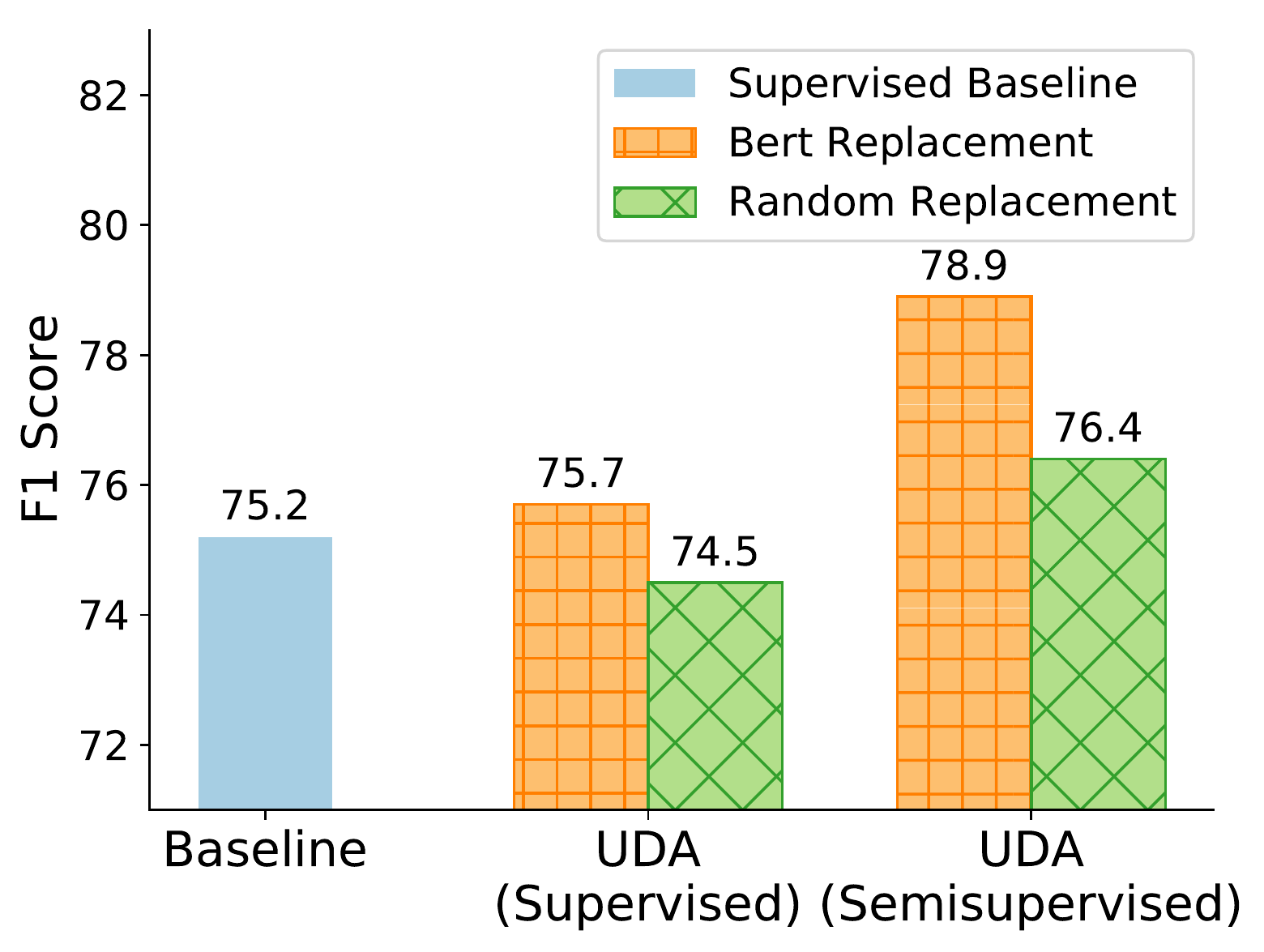}
		\caption{CoNLL}
	\end{subfigure}
    \begin{subfigure}[]{0.24\linewidth}
		\includegraphics[width=\linewidth]{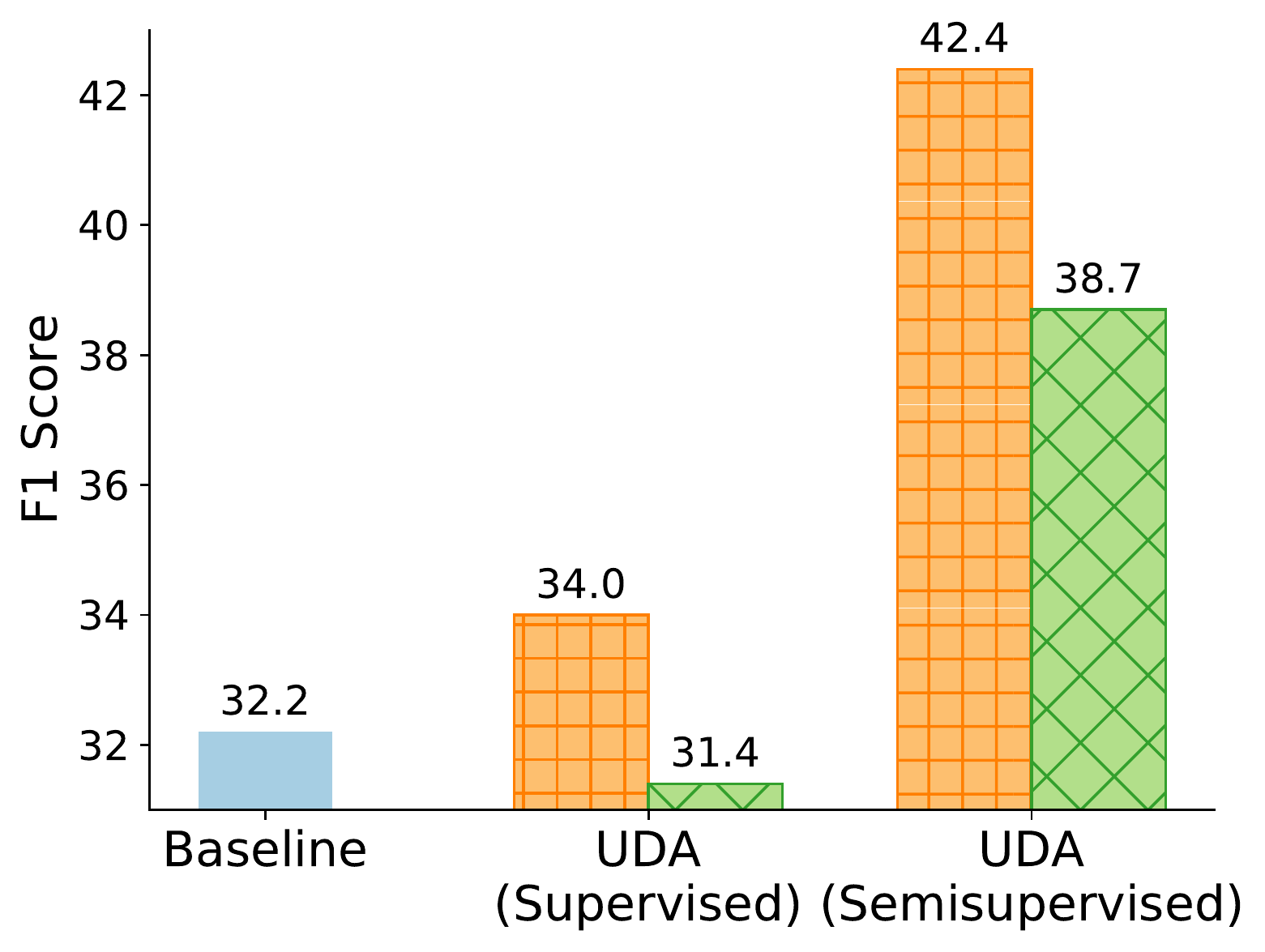}
		\caption{TAC}
	\end{subfigure}
    \begin{subfigure}[]{0.24\linewidth}
		\includegraphics[width=\linewidth]{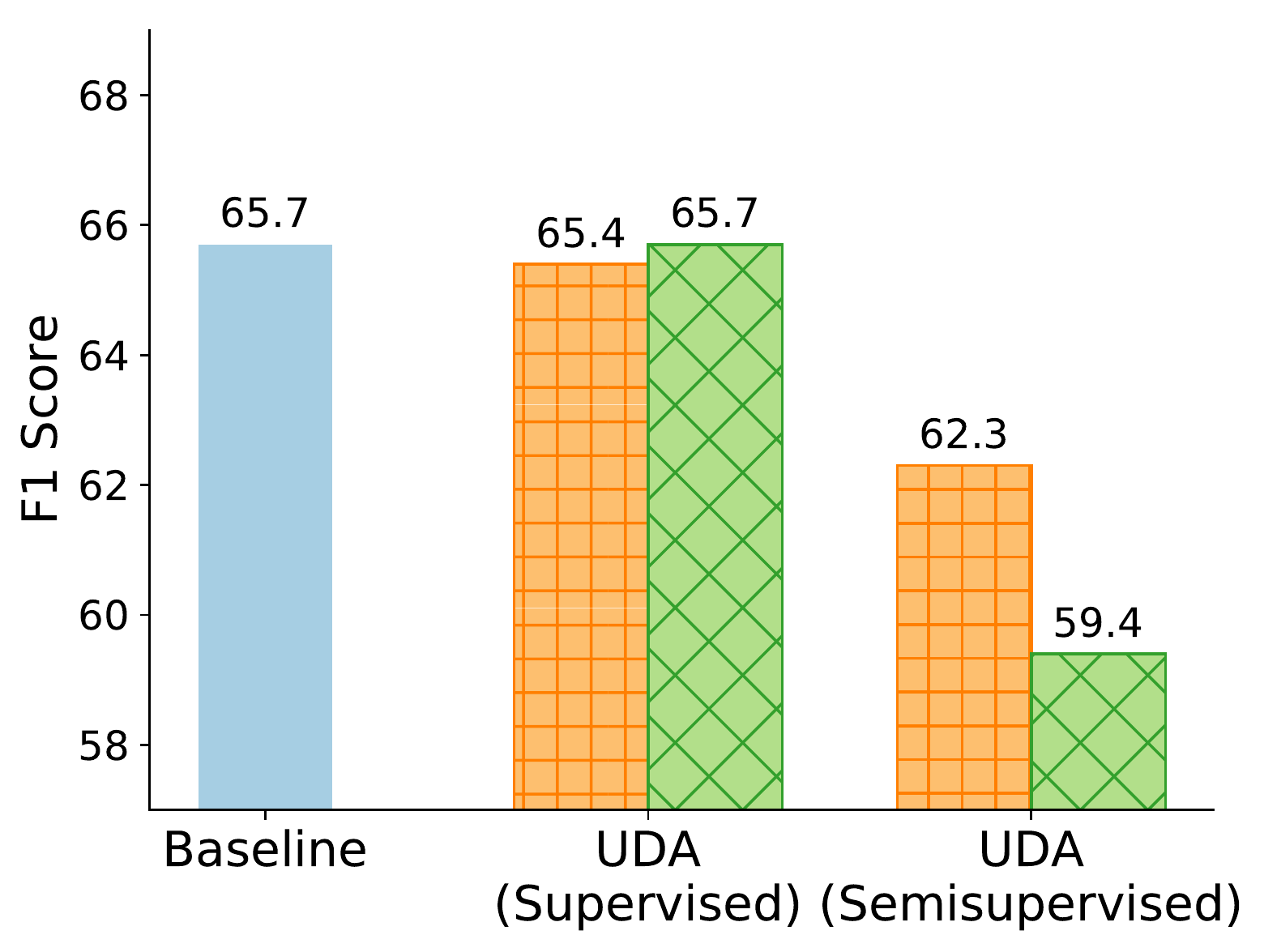}
		\caption{EBMNLP}
	\end{subfigure}
\caption{Comparison of performance achieved on sequence tagging tasks using different variants of UDA. Each bar represents the average performance across ten sets of labeled data (labeled data quantity is noted in parenthesis). The supervised baseline represents standard ML on the supervised data set only, without any consistency loss. Supervised with consistency loss represents use of consistency loss, but only over the labeled data, with the unlabeled data discarded. Semisupervised with consistency loss represents use of consistency loss over the entire dataset, both labeled and unlabeled.
}
\label{fig:seq_results}
\end{figure*}

\subsection{Sequence Tagging}

Figure \ref{fig:seq_results} presents results from sequence tagging experiments. 
BERT-based replacement provides a meaningful advantage over the supervised baseline on the CoNLL and TAC datasets.
Random replacement also offers gains, but these are smaller and less consistent.
%For these datasets, random replacement provides a less consistent and smaller advantage.
Without access to the full unlabeled dataset, random replacement results in a small decrease in performance.
With access to unlabeled data, it produces only a small benefit on CoNLL.
The gain on TAC is larger, but still smaller than that achieved using BERT-based replacement.

BERT-based replacement is more effective than random replacement for sequence tagging.
But that random replacement provides any benefits at all for such tasks is perhaps counter-intuitive, given that predictions are made at the word level for these tasks.
It is therefore likely that random replacement will lead to a change in ground truth labeling for any replaced entities.
We hypothesize that this training encourages the model to place greater weight on the context in which words appear.
This may render models more robust in being able to recognize unfamiliar entities based on the contexts in which they appear.
%Intuitively, this could be beneficial, as humans also rely on context to infer the meaning of unknown words. % bcw: I don't think we need to appeal to humans here
%BERT-based replacement is still more effective than random replacement for sequence tagging tasks, suggesting that there is still value in maintaining the ground truth labelling of the task.

UDA does not offer performance gains on the EBMNLP dataset using either augmentation strategy.
When consistency loss is applied to only the labeled data, the performance is largely unchanged.
However, when unlabeled data is incorporated, performance decreases by $3.4$ F1 when using BERT-based replacement, and $6.3$ F1 when using random replacement.
Crowdsourced (lay) workers annotated EBMNLP's training data, while doctors annotated the test data \citep{DBLP:journals/corr/abs-1806-04185}, and we speculate that this may play a role in the difference in observed performance, as we observed similar gains to those attained on CoNLL and TAC when performing exploratory studies on a development set.
It may be that UDA performs poorly on EBMNLP relative to the supervised learning baseline because relying more heavily on context is harmful when the training set annotations are noisy.
We note that the lay training set annotators consistently included more words in their labeled entity spans than the test set annotators (see Table \ref{tab:ebmnlp}).
This may indicate that the training set annotators included context words which do not truly belong to an entity class in their spans.
Encouraging the model to infer the implications of context based on these misidentified context words may be compounding that error. 

\begin{table}[]
\small
\centering
\begin{tabular}{cccc}
      & P   & I   & O \\
\hline
Train & 8.2 & 3.9 & 4.8 \\
Test  & 6.5 & 1.8 & 3.7 \\
\hline
\end{tabular}
\caption{Average length of PIO spans in words for EBMNLP's train and test sets}
\label{tab:ebmnlp}
\end{table}

Our results indicate that unlabeled data is more critical for UDA in sequence tagging than in classification.
Here, in the best case (TAC with  Replacement) we see UDA without unlabeled data achieves only $18\%$ of the performance increse 
that may ultimately be achieved by including unlabeled data.
This is lower than the worst case observed in the classification task.

\subsection{Varying Quantities of Labeled and Unlabeled Data}

\begin{figure}[htb]
    \centering
	\begin{subfigure}[]{0.49\linewidth}
		\includegraphics[width=\linewidth]{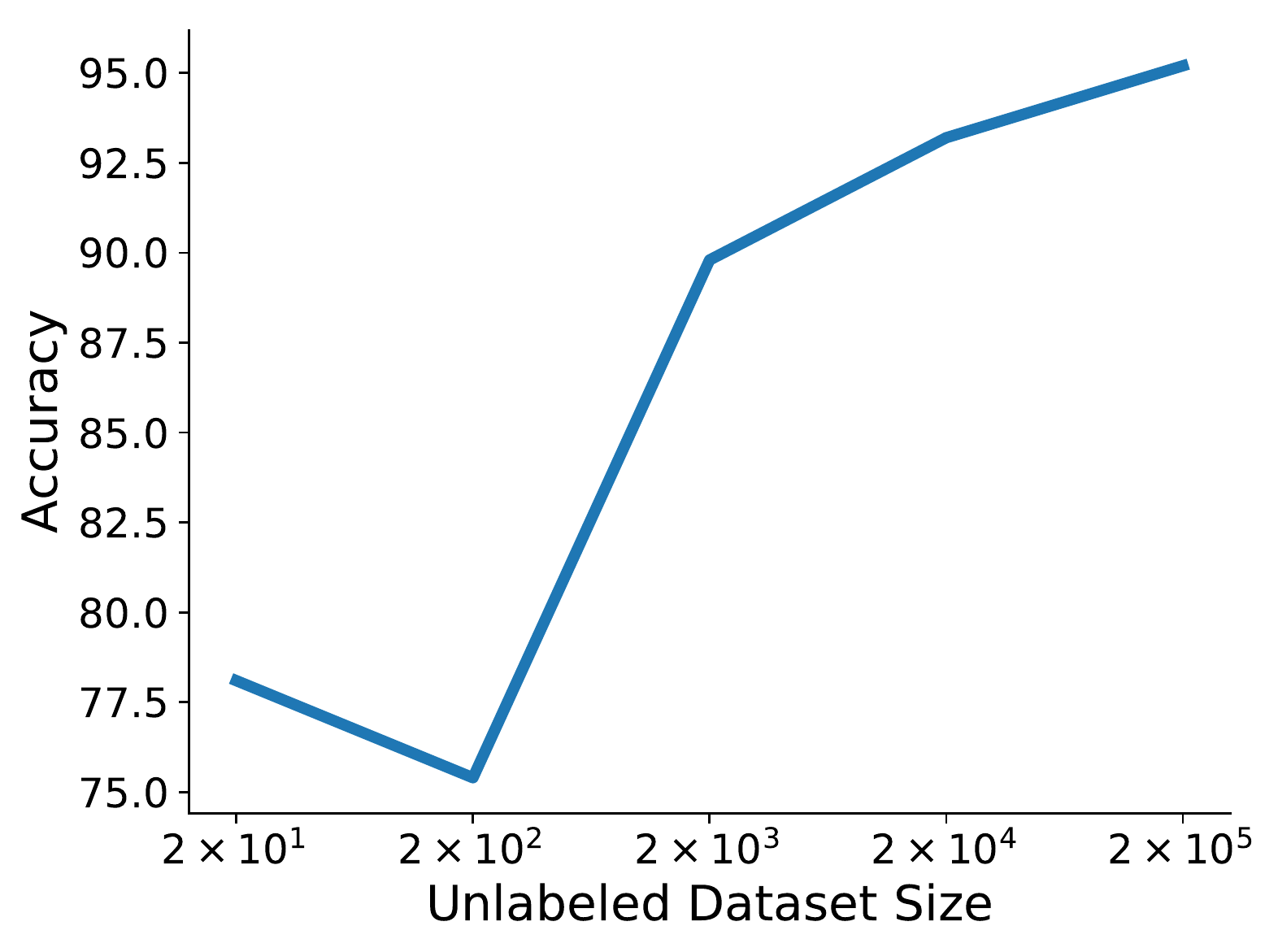}
		\caption{Yelp}
	\end{subfigure}
    \begin{subfigure}[]{0.49\linewidth}
		\includegraphics[width=\linewidth]{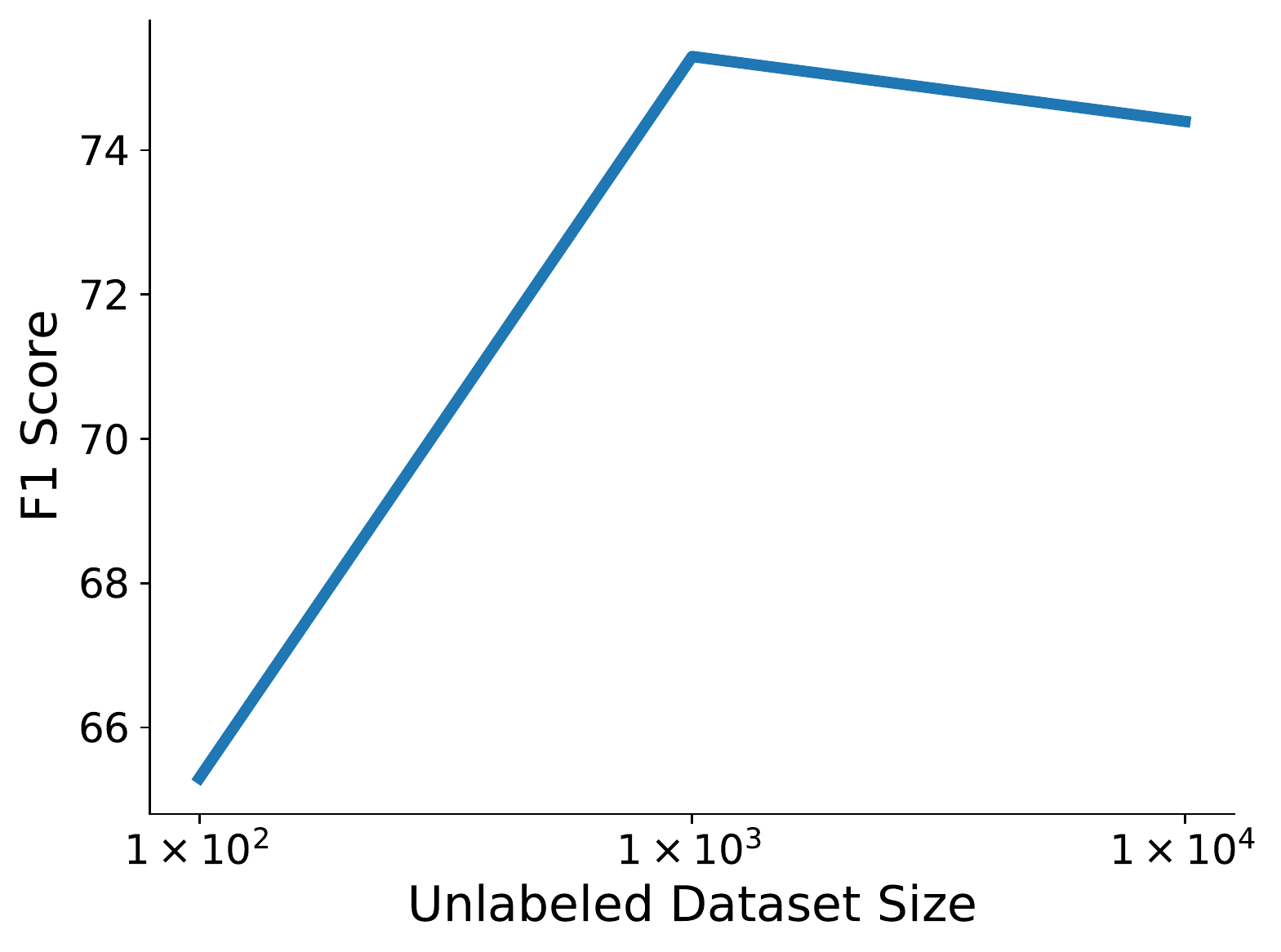}
		\caption{CoNLL}
	\end{subfigure}
\caption{Comparison of performance achieved using varying quantities of unlabeled data. Curves are averaged across 5 experiments for Yelp, and 10 for CoNLL. Each labeled dataset consists of 20 labeled examples for Yelp and 100 examples for CoNLL. Random replacement is used as the augmentation method for Yelp and BERT-based replacement is used as the augmentation method for CoNLL.
}
\label{fig:unsup_curves}
\end{figure}

\begin{figure}[htb]
    \centering
		\includegraphics[width=.85\linewidth]{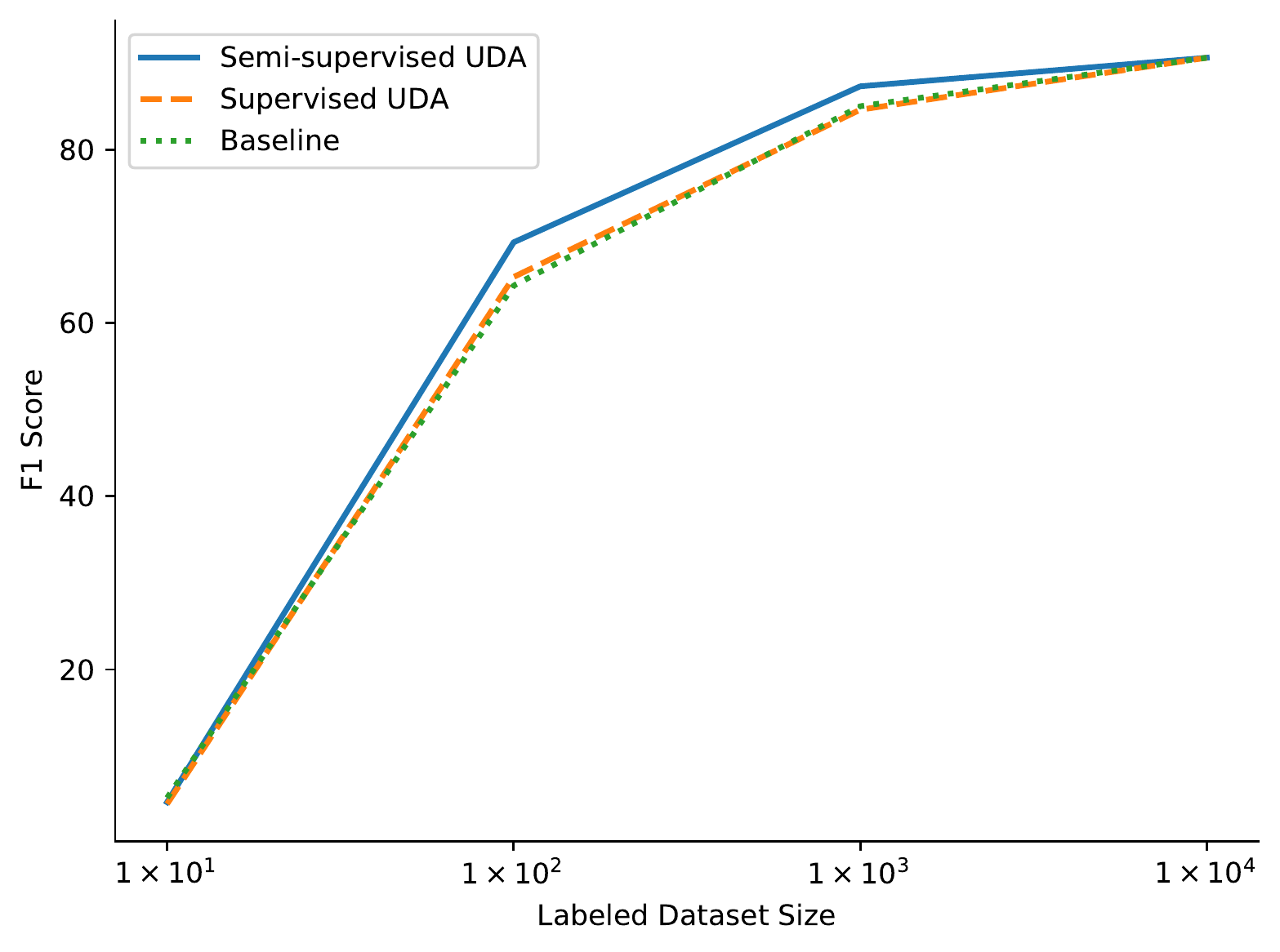}
\caption{Analysis of performance achieved using varying quantities of labeled data on the CoNLL set. The blue solid line represents the case that UDA is used and all unlabeled data is incorporated in training. The orange dashed line represents the case that UDA is used, but with only the labeled data. The green dotted line represents training without any consistency loss. Each curve is averaged across ten experiments. BERT-based replacement is used as the augmentation method.
}
\label{fig:sup_curves}
\end{figure}

We also examine the question: how much unlabeled data is necessary? Incorporating additional unlabeled data extends the training process and we hypothesize that, at some point, we will observe diminishing returns. To this end, we run experiments varying the quantity of unlabeled data used when training on the Yelp and CoNLL datasets. These results are presented in figure \ref{fig:unsup_curves}. For yelp, we begin to observe diminishing returns as we approach use of the full unlabeled set. Interestingly, on CoNLL, too high a quantity of unlabeled data appears to actually degrade performance.

We also analyze the change in performance when varying the amount of labeled data used for sequence labeling tasks. To investigate this, we train using UDA with $10$, $100$, $1000$ or $10000$ labeled sentences drawn from CoNLL. Results from these experiments are presented in Figure \ref{fig:sup_curves}. %In these experiments 
We observe consistent, modest gains from using UDA in a semi-supervised fashion, excepting the extreme ends of the curve where almost all or almost none of the data is labeled. In these two cases, we see performance with UDA and without UDA converge.

\subsection{Finetuning BERT Weights}

\begin{figure}%[htb]
	\includegraphics[width=\linewidth]{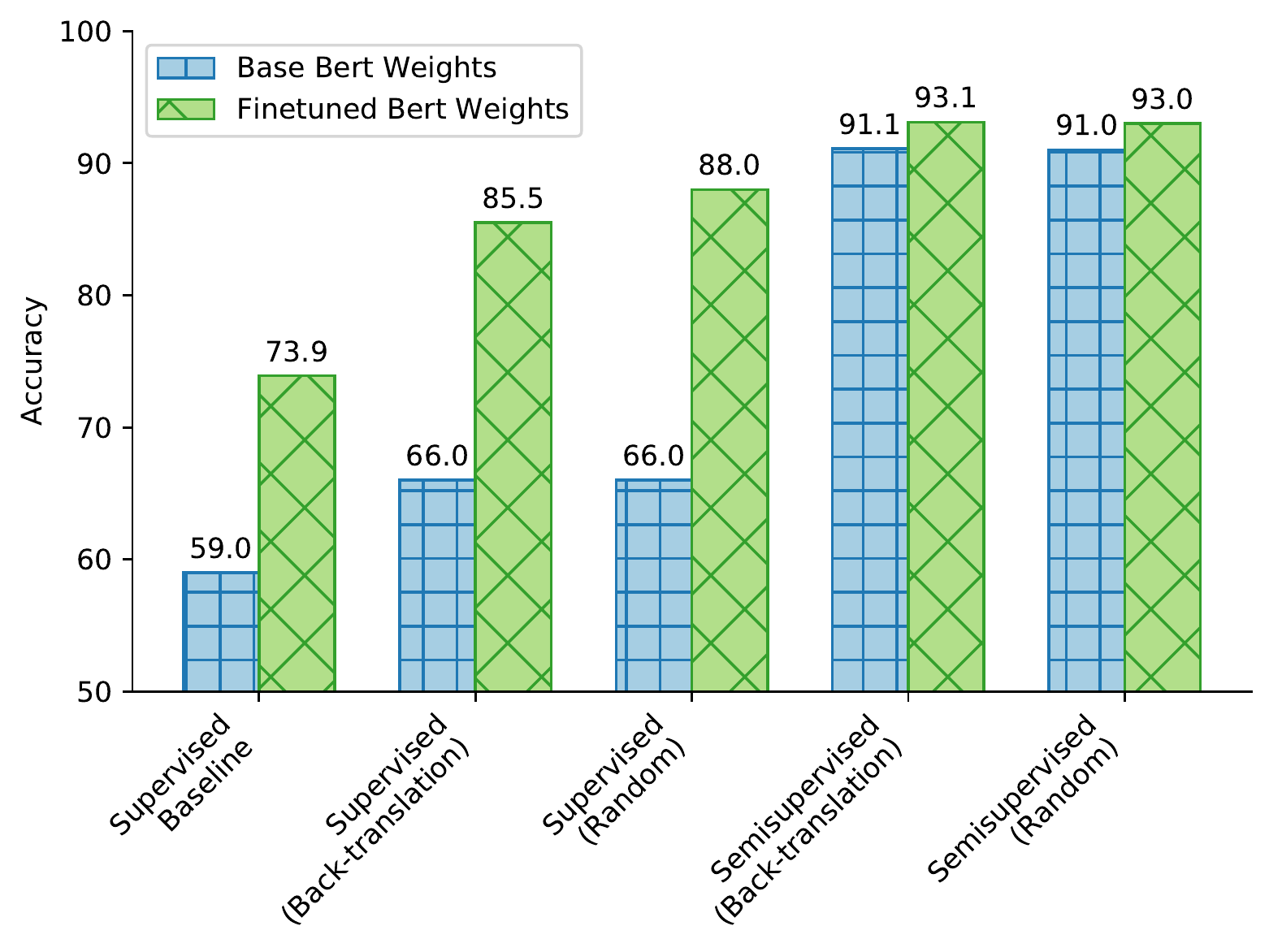} \vspace{-.5em}
\caption{Performance on IMDB using off-the-shelf pretrained BERT weights compared to BERT weights finetuned to IMDB (i.e., after continuing pretraining BERT on IMDB.}
\label{fig:finetune}
\vspace{-.25em}
\end{figure}

BERT's pretraining tasks (masked-language modeling and next
sentence prediction) already provide a method for incorporating unlabeled data \citep{devlin2018bert}.
Given this, we finetune  by training BERT's pretraining tasks on the full unlabeled datasets.
We then investigate the resulting performance on the downstream tasks, to determine whether there is still a benefit to using UDA with the full unlabeled dataset after that data has been incorporated into the model via finetuning.

Figure \ref{fig:finetune} presents results from this experiment for the IMDB dataset.
Our results show that, when BERT's weights have already been finetuned on the unlabeled data, incorporating that data again when training with UDA is less valuable.
Applying UDA using only the labeled data and random replacement allows us to realize $74\%$ of the possible performance increase when using the full unlabeled data.
This is compared to only $22\%$ when BERT has not been finetuned.

However, we still observe that performance gains do continue to accrue when unlabeled data is incorporated into UDA training, even when the BERT weights have been finetuned.
Since training with UDA is comparatively computationally inexpensive to robustly finetuning BERT, it is likely practical and advantageous to use both in concert.

\subsection{Variability of Results}

\begin{figure}%[htb]
	\includegraphics[width=\linewidth]{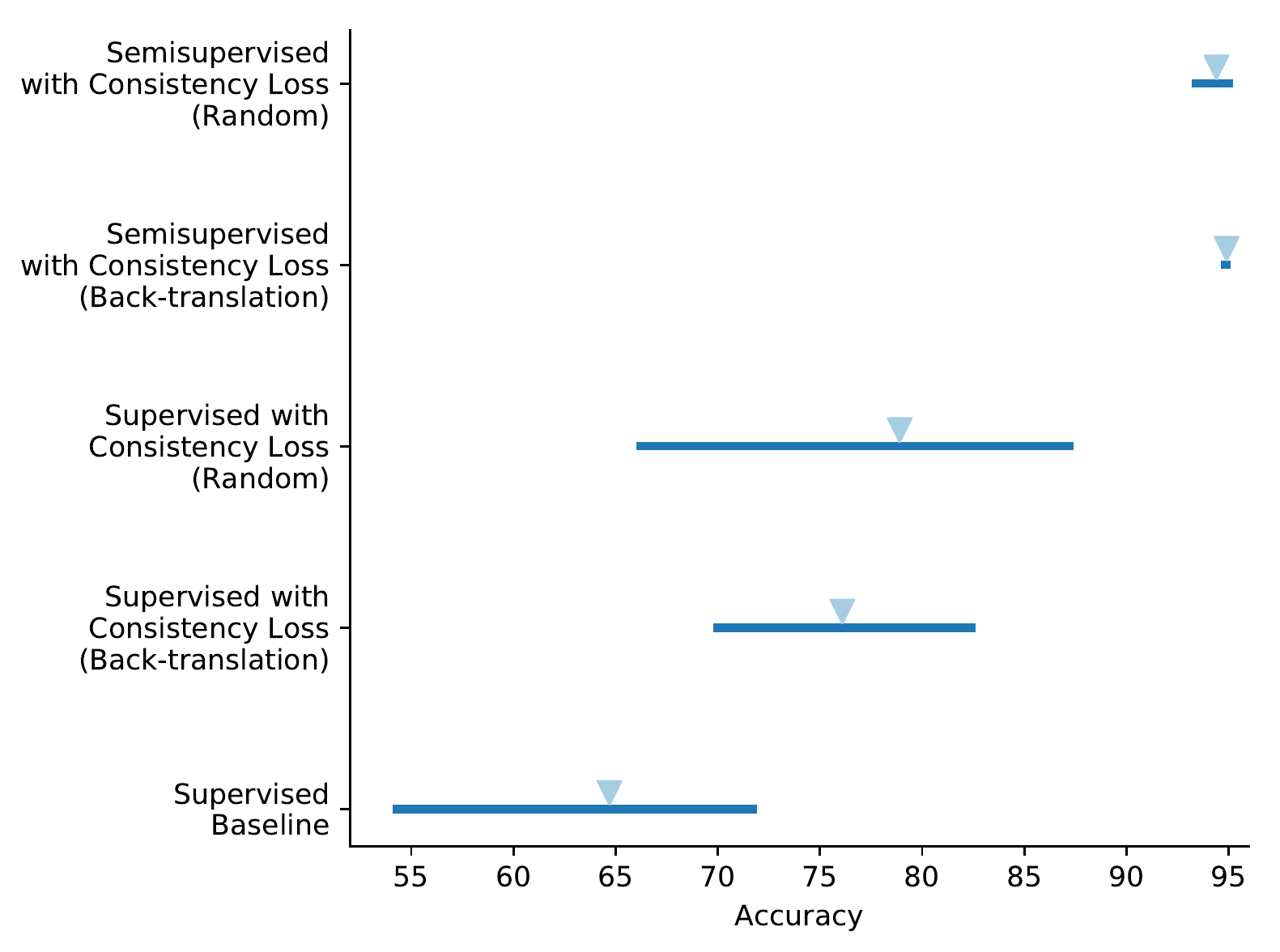}
\caption{Performance ranges on the Amazon dataset 
% across labeled set choices.
% across samples (of the labeled subset).
with spans indicating the minimum-to-maximum performance 
over 5 independent samples (of the labeled subset).
% to the maximum observed performance. 
Triangles indicate means.
% Each span is from the minimum observed performance to the maximum observed performance. Triangles indicate the mean.
}
\label{fig:ranges}
\vspace{-.5em}
\end{figure}

Throughout our experiments we observed that performance varies greatly with the choice of data to label.
Figure \ref{fig:ranges} illustrates the range of observed results on the Amazon dataset.\footnote{Similar plots for other datasets are presented in the Appendix.}
The difference between the maximum performance and the minimum performance in the supervised baseline is $17.8$ points of accuracy.
The delta for UDA using only labeled data and random word replacement is even higher, at $21.4$ points of accuracy.
This has important implications for a practitioner: While one might reasonably have an expectation of achieving high performance on average, in practice only a single labeled dataset will be constructed and used for training.
Our results show that, in a low resource classification setting, such a practitioner might actually achieve significantly lower or higher performance than expected.

This illustrates a further advantage of UDA.
When exploiting a large quantity of unlabeled data, performance not only improves, but becomes consistent across labeled dataset choices as well.
We observe similar trends across all classification datasets, with the exception of Evidence Inference with random replacement, for which the variability of results remains relatively high, even when all unlabeled data is employed.

By contrast, we do not observe such trends in the sequence tagging tasks, where UDA variant choice does not consistently effect the variability of performances across labeled set choices.

\section{Conclusions}

%In this paper we have extended the recently proposed Unsupervised Data Augmentation (UDA) method \citep{xie2019unsupervised}.
In this paper we have evaluated and extended Unsupervised Data Augmentation (UDA) in the context of NLP tasks.
We proposed and evaluated new approaches for UDA suitable to classification tasks
and extended it to sequence tagging tasks 
by imposing a consistency loss over word label distributions.
We showed that naive data augmentation methods 
may be just as effective as the complex,
model based augmentations currently in use,
and that performance improvements may still be attained 
even in the absence of any unlabeled data.

More specifically, we proposed a simple, 
effective augmentation method: 
randomly replace words with other words.
The replacement word may be selected either uniformly at random, 
or by using BERT \citep{devlin2018bert} as a masked language model 
to induce a probability distribution over tokens.% candidates.
We found that the former method is effective for classification tasks, 
and the latter %is effective 
for sequence tagging.
We further investigated the practicality of using UDA without unlabeled data, 
applying a consistency loss only to a small labeled dataset.
We experimentally evaluated various augmentation strategies and settings 
on four classification datasets and three sequence tagging datasets.

We found reliable performance increases on all four classification datasets.
For classification, we found that random word replacement is as effective as---and sometimes more effective than---back-translation,
which is what has been proposed in prior work \citet{xie2019unsupervised}.
In particular, random word-replacement 
is effective on our scientific classification task, 
where back-translation is ineffective, 
perhaps due to its reliance on machine translation models 
not trained with scientific literature.

We found that both random replacement and BERT-based replacement 
are effective on two out of three sequence tagging tasks, 
with BERT-based replacement that we have proposed 
consistently outperforming random replacement.
On the third sequence tagging dataset,
we observed a degradation of performance when using any variety of UDA,
which we hypothesize owes to the noisy annotation of this training set.
%is due to noisy annotation of the training set.

%We further found that 
We found that UDA may produce meaningful increases in performance 
even when unlabeled data is not available, 
particularly if the weights have already been finetuned to the task. % bcw: careful with `finetuned' hyphen/no hyphen -- i dont care either way but want to be consistent!
The magnitude of this increase depends upon the task, 
and may be relatively large (as in the case of the Amazon dataset)
or relatively small (as in the CoNLL dataset).
In general this approach is more effective 
for classification tasks than sequence tagging tasks.

To summarize our findings: UDA is effective in low-supervision natural language tasks,
even when used with naive augmentation methods and without unlabeled data.

\bibliography{eacl2021}
\bibliographystyle{acl_natbib}

\end{document}